\documentclass[letterpaper, 10 pt, conference, fleqn]{ieeeconf}  

\IEEEoverridecommandlockouts                              

\usepackage{acro} 
\usepackage{gensymb}
\usepackage{cite}
\usepackage{float}
\usepackage{bm}
\usepackage{gensymb}
\usepackage[binary-units,product-units=power]{siunitx}
\usepackage{graphicx}

\usepackage{mathtools}
\usepackage{amsmath}
\usepackage{amssymb}
\usepackage{amsfonts}
\usepackage[mode=buildnew]{standalone}
\usepackage{booktabs} 
\usepackage[caption=false,font=footnotesize]{subfig}
\usepackage{nccmath}

\DeclareMathOperator*{\subjectto}{subj.\:to}

\DeclareAcronym{CHA}{
  short = CHA,
  long  = confirmed hazardous area
}

\DeclareAcronym{DDPS}{
  short = DDPS,
  long  = Federal Department of Defence\, Civil Protection and Sport
}
\DeclareAcronym{DOF}{
  short = DOF,
  long  = degrees of freedom
}
\DeclareAcronym{GNSS}{
  short = GNSS,
  long  = global navigation satellite system
}
\DeclareAcronym{GPR}{
  short = GPR,
  long  = ground penetrating radar
}
\DeclareAcronym{IMU}{
  short = IMU,
  long  = inertial measurement unit
}
\DeclareAcronym{LiDAR}{
  short = LiDAR,
  long  = light detection and ranging
}
\DeclareAcronym{UXO}{
  short = UXO,
  long  = unexploded ordnance
}
\DeclareAcronym{ERW}{
  short = ERW,
  long  = explosive remnants of war,
  short-indefinite  = an,
  long-indefinite  = an
}
\DeclareAcronym{EKF}{
  short = EKF,
  long  = extended Kalman filter
}
\DeclareAcronym{LIO}{
  short = LIO,
  long  = LiDAR inertial odometry
}
\DeclareAcronym{MAV}{
  short = MAV,
  long  = micro aerial vehicle,
  short-indefinite  = an
}

\DeclareAcronym{SHA}{
  short = SHA,
  long  = suspected hazardous area
}
\DeclareAcronym{FOV}{
  short = FOV,
  long  = field of view
}

\DeclareAcronym{SDF}{
  short = SDF,
  long  = signed distance field,
  short-indefinite  = an
}

\DeclareAcronym{NCCR}{
  short = NCCR,
  long  = National Center of Competence in Research,
  short-indefinite = am,
  long-indefinite = a
}
\DeclareMathOperator*{\argmin}{arg\,min}
\DeclareMathOperator*{\argmax}{arg\,max}

\title{\LARGE \bf
Resilient Terrain Navigation with a 5~\acs{DOF} Metal Detector Drone
}

\author{Patrick Pfreundschuh, Rik Bähnemann, Tim Kazik, Thomas Mantel, Roland Siegwart, and Olov Andersson
\thanks{This work was supported by Armasuisse W+T, Urs Endress Foundation's Project FindMine, a Wallenberg Foundation and WASP Postdoctoral Scholarship, and Swiss National Science Foundation's \acs{NCCR} DFab P3.}
\thanks{Authors are with Autonomous Systems Lab, ETH Zurich, e-mail: \texttt{\small \{patripfr, brik, tkazik, mantelt, rsiegwart, nandersson\}@ethz.ch}}%
}

\begin{document}

\maketitle
\thispagestyle{empty}
\pagestyle{empty}

\begin{abstract}

\Acp{MAV} hold the potential for performing autonomous and contactless land surveys for the detection of landmines and \ac{ERW}. Metal detectors are the standard detection tool but must be operated close to and parallel to the terrain. A successful combination of \acp{MAV} with metal detectors has not been presented yet, as it requires advanced flight capabilities. To this end, we present an autonomous system to survey challenging undulated terrain using a metal detector mounted on a \si{5}~\ac{DOF} \ac{MAV}. Based on an online estimate of the terrain, our receding-horizon planner efficiently covers the area, aligning the detector to the surface while considering the kinematic and visibility constraints of the platform. As the survey requires resilient and accurate localization in diverse terrain, we also propose a factor graph-based online fusion of GNSS, IMU, and LiDAR measurements. We validate the robustness of the solution to individual sensor degeneracy by flying under the canopy of trees and over featureless fields. A simulated ablation study shows that the proposed planner reduces coverage duration and improves trajectory smoothness. 
Real-world flight experiments showcase autonomous mapping of buried metallic objects in undulated and obstructed terrain. 

\end{abstract}

\section{INTRODUCTION}

Worldwide, landmines and \ac{ERW} caused more than \SI{7000}{} casualties in 2020. Member states of the Mine Ban Treaty reported that approximately \SI{12000}{\kilo\metre\squared} are contaminated, making vast areas uninhabitable~\cite{landminemonitor2021}.
Reducing \aclp{SHA} is challenging, as those areas are inaccessible. Additionally, detecting and mapping potential contamination requires expert training. As land clearance is expensive, it should be limited to minimal \aclp{CHA}~\cite{imas0711}.

Recently, \acp{MAV} with various sensors have been investigated to accelerate land release. 
Drones can access any terrain without putting a person at risk and automatically collect geo-referenced data. In contrast to ground-based platforms, \acp{MAV} can inspect a surface contactless, relaxing traversability limitations~\cite{balta2013terrain} and eliminating the risk of accidental detonations.
The essential sensor for humanitarian demining is the metal detector. 

However, implementing a metal detector on a drone has failed so far because of the advanced flight capabilities necessary to perform practical surveys.
Detection performance highly depends on the sensor head alignment with the soil surface~\cite{de2003humanitarian}.
Thus, the drone needs to maintain distance while aligning the sensor parallel with the surface.
This requires a customized mechanical platform concept and online perception to navigate resiliently in unknown, obstructed, undulated, outdoor terrain. 

\begin{figure}
    \centering
\includegraphics[width=\linewidth]{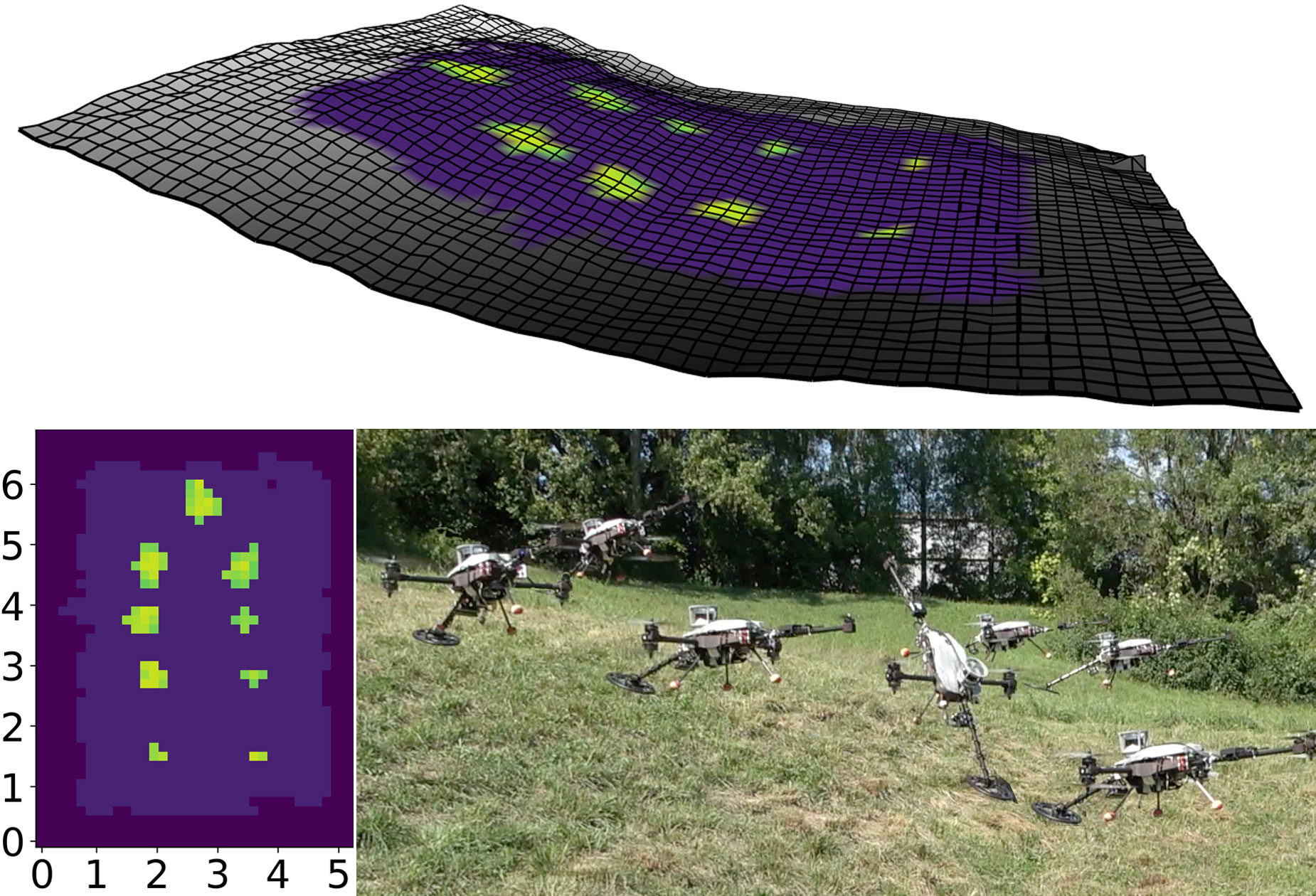}
    \vspace{-6mm}
    \caption{\textit{Top}: Metal detector signals visualized on a 2.5D terrain map. Colored cells were covered by the sensor. Brighter colors correspond to stronger signal responses from buried objects. \textit{Bottom left}: Top-down 2D detection map. \textit{Bottom right}: The system in action on undulated terrain.}
    \label{fig:undulated_heat}
    \vspace{-4.5mm}
\end{figure}

This work tackles this challenging surface tracking problem with a novel flying metal detector design.
The system consists of a \si{5}~\ac{DOF} drone with a rigidly attached sensor.
The drone can maneuver in all three translational directions and independently change its heading and pitch to align the sensor with the surface (Fig.~\ref{fig:undulated_heat}).
Simply aligning the sensor according to the surface normal is not ideal, as small changes in the normal can cause large motions of the platform as illustrated in Fig.~\ref{fig:sim}. As this is inefficient, we propose a new local trajectory planner that trades off optimal alignment with minimal turning efforts and collision avoidance.

During the survey, the system maps the terrain with \acs{LiDAR}.
Precise surface tracking and complete coverage require accurate, smooth localization to keep the sensor close to the ground without touching it. The margins for inaccuracies are small, as they could result in a crash or invalid survey results, but \ac{ERW} are often close to natural or human-made structures that cause \acs{GNSS} outages or in feature-less areas which cause \ac{LIO} drift. We thus fuse these complementary modalities for resilient localization.

The main contributions of our work are:
\begin{itemize}
    \item An autonomous system for a flying metal detector based on a \si{5}~\ac{DOF} \ac{MAV}.  
    \item A receding-horizon trajectory planner that optimizes terrain alignment in real-time while considering drone maneuverability and online map updates.
    \item Online factor graph \acs{GNSS}, \acs{IMU}, and \acs{LiDAR} localization that is resilient to individual sensor degeneracy.
\end{itemize}

We show in simulation that our receding-horizon planning reduces coverage duration while maintaining sensor alignment. Real-world experiments show autonomous surveying of undulated and obstructed terrain and consistent localization even under the loss of \acs{GNSS} or drift in \ac{LIO}.

\section{Related Work}
\subsection{Flying Robotic \ac{ERW} Detection}
Several works towards \ac{ERW} detection using aerial platforms exist. Some systems use ground penetrating radars \cite{bahnemann_under_2022, lopez2022unmanned, sipos_2017} or cameras \cite{baur2020applying}. While these sensors can survey large areas from relatively far distances, standardization, and target detection are still being investigated. Yoon et al. \cite{yoo_application_2021} use a magnetometer but require special data processing to detect small objects.
The most accepted sensor for deminers is the metal detector. Due to the limited sensing range of metal detectors, the sensor head must remain close and parallel to the surface for a strong sensor response~\cite{de2003humanitarian}. Underactuated \acp{MAV} have coupled translational and rotational dynamics, which makes surface alignment challenging. The Mine Kafon Drone~\cite{mine_kafon} proposed to solve this by mounting the metal detector on a gimbal but has not reported results yet. Our \si{5}~\ac{DOF} \ac{MAV} with a rigidly mounted metal detector avoids the necessity of a gimbal as the vehicle pose directly controls the sensor orientation. Furthermore, our system is the first to perceive the environment online and avoid obstacles. 

\subsection{Resilient Online State Estimation with \acs{GNSS} and \ac{LIO}}
Both loosely- and tightly-coupled sensor fusion leverage the complementary nature of \acs{GNSS} and \ac{LIO}. The authors in \cite{nubert_graph-based_2022} present a loosely-coupled approach that deploys a dual-factor graph to fuse the different modalities. In contrast, \cite{li_gil_2021} uses an \ac{EKF} to perform a tightly-coupled fusion of \ac{LiDAR}, inertial, and \acs{GNSS} measurements. LIO-SAM~\cite{shan_lio-sam_2020} uses a factor graph~\cite{factor_graphs} to tightly fuse \ac{IMU} and \acs{GNSS} with scan-matching \ac{LIO}. These approaches assume that the transformation between the local odometry and the \acs{GNSS} frame is directly measured using a second GNSS receiver \cite{nubert_graph-based_2022} or estimated in advance \cite{shan_lio-sam_2020, li_gil_2021} with an external pipeline that requires an initialization motion.
Inspired by visual-inertial GNSS fusion \cite{mascaro_gomsf_2018, boche_visual-inertial_2022, lee_intermittent_2020}, we formulate this transformation as an extrinsic online calibration. In contrast to the LIO-GNSS fusion frameworks above, we fuse the GNSS measurements into the odometry frame without requiring prior knowledge about the heading offset between the reference frames. Additionally, we define it as a transformation that varies over time to compensate for drift accumulating from \ac{LIO} over periods of \acs{GNSS} dropouts. 

\subsection{Trajectory Planning on Surfaces}
Finding the optimal flight trajectory for surface coverage in unknown 3D environments under visibility constraints is a challenging planning problem. Since the exact solution is computationally intractable, this work assumes a given coverage path for a region of interest. We then attempt to find a \si{5}~\ac{DOF} flight trajectory that accurately tracks the terrain surface to ensure good alignment for \ac{ERW} detection, while, in real-time, considering terrain visibility, speed, smoothness, and avoiding collisions. Earlier work addresses either planning on manifolds without considering coverage or coverage planning that either does not consider \ac{MAV} kinematics or runs only offline. Manifold planning approaches include \cite{lu2021model}, which uses a model-predictive control approach, and \cite{pantic2021mesh}, which proposes a reactive planner based on Riemannian motion policies. Visibility constraints are also included in \cite{watterson2020trajectory} but do not consider coverage or run in real-time. On the other hand, \cite{davis2016c} proposes a belief-space approach to optimize the coverage of free-flight \ac{MAV} trajectories, but it is vulnerable to local minima and does not run onboard. Finally, \cite{choi2018three} plans coverage patterns high above 3D terrain for remote sensing but does not consider \ac{MAV} kinematics or appear suitable for real-time flight.

\section{Platform Setup}
\begin{figure}[t!]
    \centering
    \includegraphics[width=0.75\linewidth]{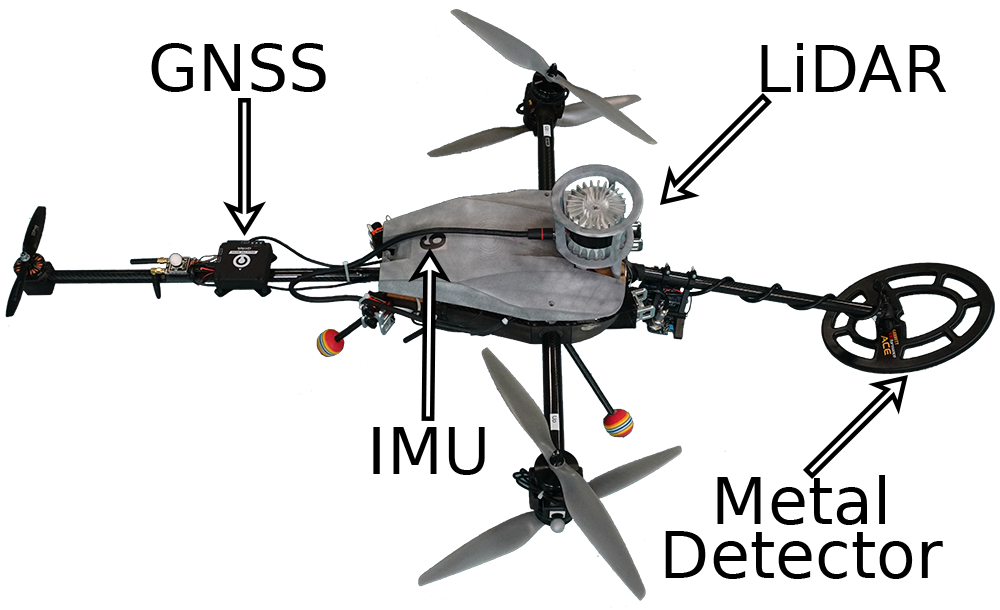}
     \vspace{-2mm}
    \caption{The Voliro-T tricopter with a custom payload for autonomous navigation and metal detection.}
    \vspace{-5mm}
    \label{fig:voliro_t}
\end{figure}
We built our system on top of a tricopter from Voliro~AG. The platform, shown in Fig.~\ref{fig:voliro_t}, has fully actuated front propellers, allowing position control independent of the body pitch angle. The additional actuation also improves stable flight near surfaces where disturbances from propeller wash occur.
We equipped the platform with an Ouster OS0-128 \ac{LiDAR} with a \SI{90}{\degree} vertical \ac{FOV}, allowing it to observe close terrain. The rear-facing \ac{FOV} is partially obscured by the body. We also use an u-blox ZED-F9P multi-band RTK-GNSS receiver. A Bosch BMI085 IMU provides inertial measurements.
Finally, a Garret ACE300i metal detector with \SI{0.25}{m} major axis length is rigidly attached at the front. A custom microcontroller interface captures the digital output from the metal detector and sends it to the vehicle's onboard computer. The software stack runs on a single-board PC with an AMD 4800U x86 CPU. 

\section{Resilient \acs{GNSS}, \acs{IMU} and \acs{LiDAR} Fusion}\label{gnss_lio}
This section describes our GNSS and LIO fusion. Existing approaches \cite{nubert_graph-based_2022,
shan_lio-sam_2020, li_gil_2021} fuse the odometry into the GNSS frame, which requires prior knowledge of the yaw between the two reference frames. Our system directly operates in the LIO frame and estimates the alignment between the coordinate frames online. Thus our system is immediately deployable without prior initialization motion or sensing.

\subsection{Notation and Problem Definition}
We define a fixed gravity-aligned odometry frame $\mathbf{O}$ with origin at the initial pose of the IMU. The \acs{GNSS} receiver measures position with respect to an east-north-up gravity-aligned inertial frame $\mathbf{I}$, with origin at the position of the first measurement.
The IMU frame serves as the robot body frame $\mathbf{B}$. The detector frame $\mathbf{D}$ is placed in the center of the metal detector and aligned with the two axes of its elliptical coil. 
We define the robot's state as
$$\mathbf{x} = [\mathbf{R}_{OB}, {}_{O}\mathbf{p}_{B}, {}_{O}\mathbf{v}_{B}, {}_{B}\mathbf{b}^{a}, {}_{B}\mathbf{b}^{g}],$$
where $\mathbf{R} \in \mathit{SO}(3)$  and $\mathbf{p} \in \mathbb{R}^3$ represent the body orientation and position.
$\mathbf{v}_B \in \mathbb{R}^{3}$ is its linear velocity, while $\mathbf{b}^{a} \in \mathbb{R}^3$ and $\mathbf{b}^{g} \in \mathbb{R}^3$ indicate the accelerometer and gyro biases of the \ac{IMU}.
The transformation from the inertial to the odometry frame is
$\mathbf{T}_{OI} \in \mathit{SE}(3)$. The relative position ${}_{B}\mathbf{p}_{P}$ between the GNSS receiver and the \ac{IMU} is known. $t^i, t^j, t^k$ are IMU, LiDAR, and GNSS measurements timestamps.

\begin{figure}[t!]
    \centering
    \includegraphics[width=\linewidth]{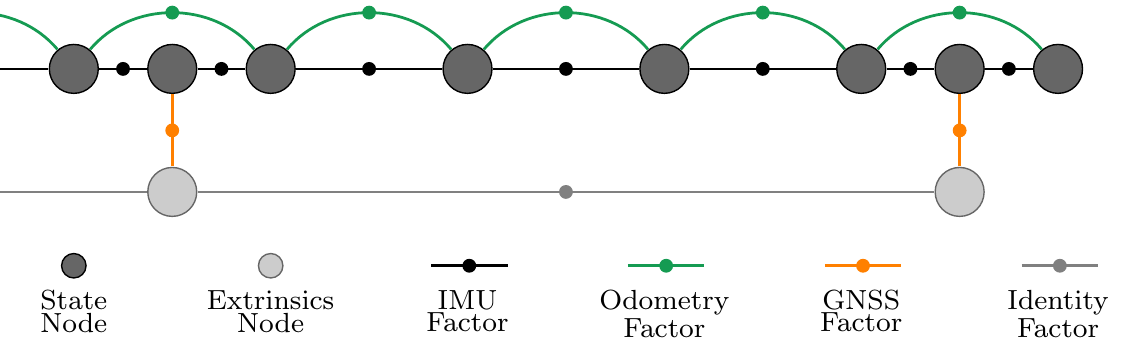}
    \vspace{-7mm}
    \caption{The factor graph design. State nodes are connected through odometry and IMU factors, while extrinsic nodes are connected through identity factors. GNSS factors connect state and extrinsics nodes.}
    \label{fig:graph}
    \vspace{-5mm}
\end{figure}

Given the set of \ac{LIO} estimates, \acs{GNSS}, and \ac{IMU} measurements, the objective is to calculate the maximum a posteriori estimate for a set of states and extrinsics. We formulate this inference problem using the factor graph depicted in Fig.~\ref{fig:graph}. We use a fixed window-length incremental smoothing and mapping (iSAM2)~\cite{isam2} solver to optimize the factor graph. In the following section, we present the individual factors.

\subsection{IMU Factor}
Pre-integrated \ac{IMU} factors summarize the \ac{IMU} measurements between two nodes into a single factor~\cite{imu_pre}.
We introduce nodes at \acs{GNSS} and \ac{LiDAR} timestamps, which arrive at a lower frequency. The \ac{IMU} integration additionally serves as a high-frequency state estimate for the flight controller~\cite{indelman2013information}. Initially, the vehicle is static to estimate an initial value for the \ac{IMU} biases and gravity alignment.

\subsection{Odometry Factor}
As a LIO pipeline, we use FAST-LIO2 \cite{fastlio}, which operates on points directly. Raw points are favorable in environments with little geometric structure, as they can represent subtle features of the scene. FAST-LIO2 estimates a 6 \ac{DOF} pose $\mathbf{T}_{OB}^j$ at each $\mathbf{t}^j$. 
We extract the relative transformation for the odometry factor as
$$\Delta\mathbf{T}^{j,j+1} = {(\mathbf{T}_{OB}^j})^{-1}\mathbf{T}_{OB}^{j+1}.$$
We base the covariance estimation on the constraints of the registration problem, similar to \cite{zhang_degeneracy}, instead of assuming a fixed covariance as in \cite{nubert_graph-based_2022,shan_lio-sam_2020}. It computes as
$$\mathbf{\Sigma}_{pos} = \mathbf{H}_{pos}=(\mathbf{J}_{pos}^T \mathbf{J}_{pos} )^{-1},    \mathbf{\Sigma}_{rot} = \mathbf{H}_{rot}=(\mathbf{J}_{rot}^T \mathbf{J}_{rot} )^{-1},$$ 
where $\mathbf{J}_{pos}$ and $\mathbf{J}_{rot}$ correspond to the translational and rotational components of the measurement Jacobian in \cite{fastlio}. A tuning constant scales the covariance magnitude.

\subsection{Position Factor}
The \acs{GNSS} receiver measures the position with respect to the inertial frame $\mathbf{I}$. The measurement influences the odometry via the extrinsics $\mathbf{T}_{OI}$.
The extrinsics' translation, roll, and pitch are directly observable, while yaw is only observable after the platform has moved sufficiently. We treat this transformation as a variable estimated in the factor graph. Position factors update the state and the extrinsics and use the covariance provided by the \acs{GNSS} receiver. Their measurement model is
$${}_{I}\mathbf{p}_{P} = (\mathbf{T}_{OI})^{-1} \cdot \mathbf{T}_{OB} \cdot {}_{B}\mathbf{p}_{P}.$$

We initialize the extrinsics with a high covariance on the yaw. Until the covariance of the yaw of the extrinsics is below a threshold, we artificially increase the \acs{GNSS} measurement covariance. Hence, the GNSS measurements have negligible impact on the state during heading estimation. Once the covariance of the extrinsic yaw estimate decreases, the GNSS measurements increase their impact on the state.

\section{Trajectory Planner for Surface Tracking}
The metal detector head has to be parallel to the ground surface to ensure reliable detection of buried objects. A fundamental part of the planning algorithm is an accurate local terrain representation. We use the 2.5D grid map framework~\cite{grid_map} to fuse the LiDAR measurements into a map. The map also provides surface normals for alignment and \iac{SDF} for collision avoidance. Second, we assume a 2D boustrophedon coverage path~\cite{bahnemann2021revisiting} given to serve as a global reference. The goal of the trajectory planner is to align the metal detector to the surface while safely tracking the reference.

Theoretically, our actuated \si{5}~\ac{DOF} drone can align with any surface by a combination of yaw $\psi(t)$ and pitch $\theta(t)$. However, alternating surface normals result in many inefficient heading changes over uneven terrain, as seen in Fig.~\ref{fig:sim}. Furthermore, uncontrolled yaw can cause the LiDAR to miss the terrain on the path ahead.
Therefore, we need to plan a trajectory over position and orientation, $(\mathbf{p}(t), \mathbf{R}(t))$, that minimizes control effort (i.e. turning) to ensure smooth and efficient trajectories that satisfy ground alignment constraints, the kinematic limits for the \ac{MAV}, and LiDAR visibility constraints to prevent flying into unobserved space.

\begin{figure}[t!]
    \centering
    \includegraphics[width=.35\textwidth]{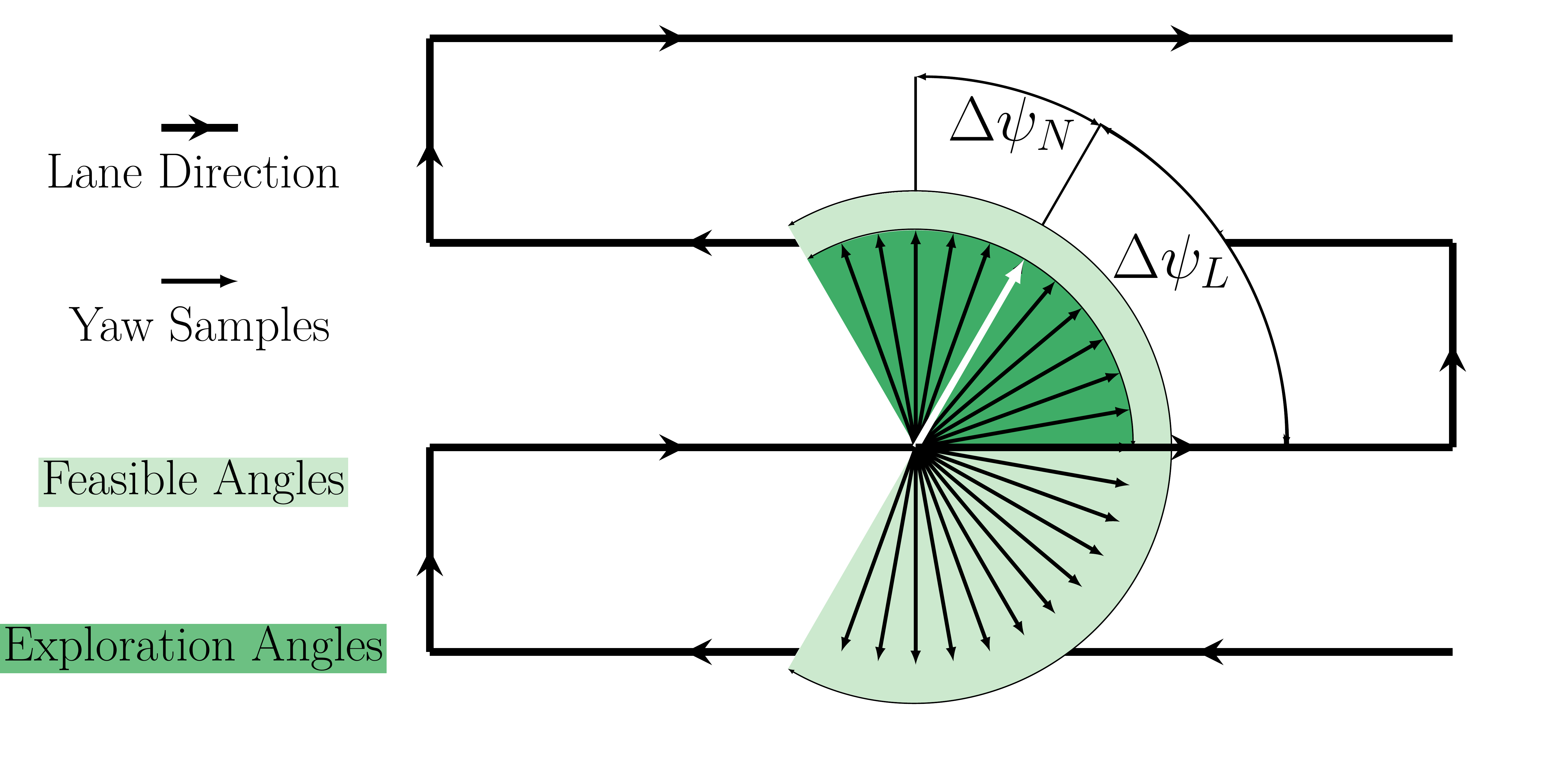}
    \includegraphics[width=.35\textwidth]{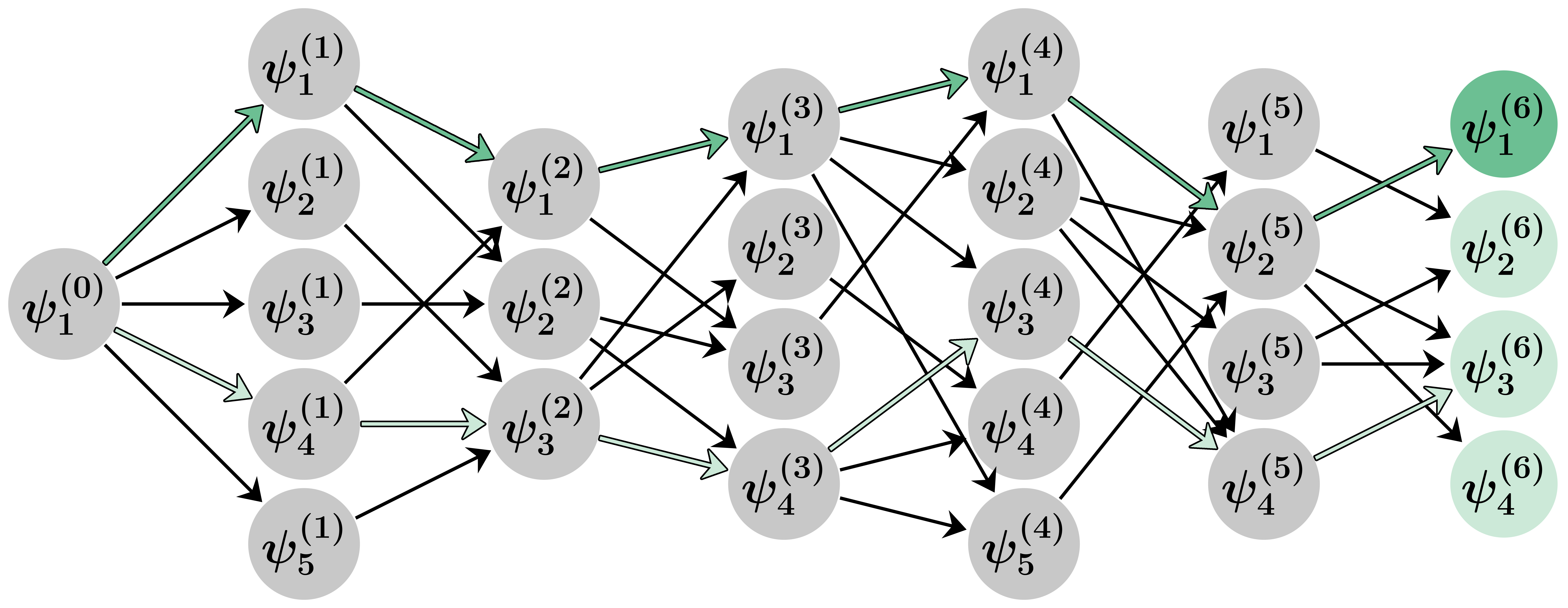}
    \vspace{-4mm}
    \caption{\textit{Top:} Illustration of the reduced planning problem. A state lattice is constructed from yaw samples along the path, which have to satisfy sensor visibility constraints (light green) of $\pm$ \SI{120}{\degree} around the direction of flight. Dark green samples are preferred as they ensure visibility of the next lane. \textit{Bottom:} Illustration of the multitree. The shortest path is indicated in light green, but as solutions in the preferred sector exist, our algorithm selects the shortest path that ends in a preferred state, as shown in dark green.
    }
    \label{fig:corridor}
    \vspace{-4mm}
\end{figure}

To transform this complex optimization problem into a real-time solution, we first show how to simplify it for a \si{5}~\ac{DOF} \ac{MAV} without obstacles before adding collision avoidance. 
Since the translational dynamics of the vectored thrust \ac{MAV} are fast~\cite{watson2021dry} compared to the maximum speed of \SI{1}{m/s} allowed for reliable detection with a metal detector \cite{user_man}, we can assume that the coverage path can be tracked in position and just optimize rotations along the path. 

\begin{figure}[t!]
    \subfloat[Angle definitions]{
        \includegraphics[width=0.35\linewidth]{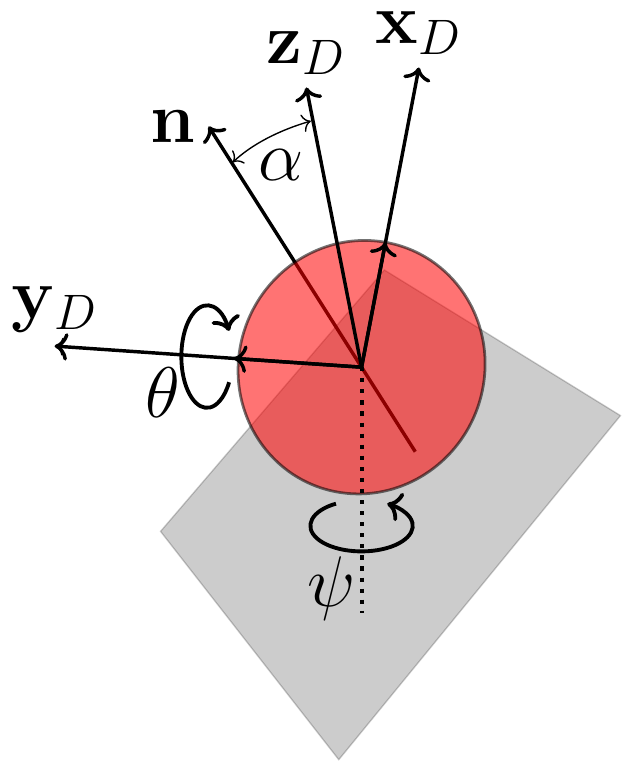}
        \label{fig:alignment}}
    \hspace*{\fill} 
    \subfloat[Obstacle avoidance]{
        \includegraphics[width=0.35\linewidth]{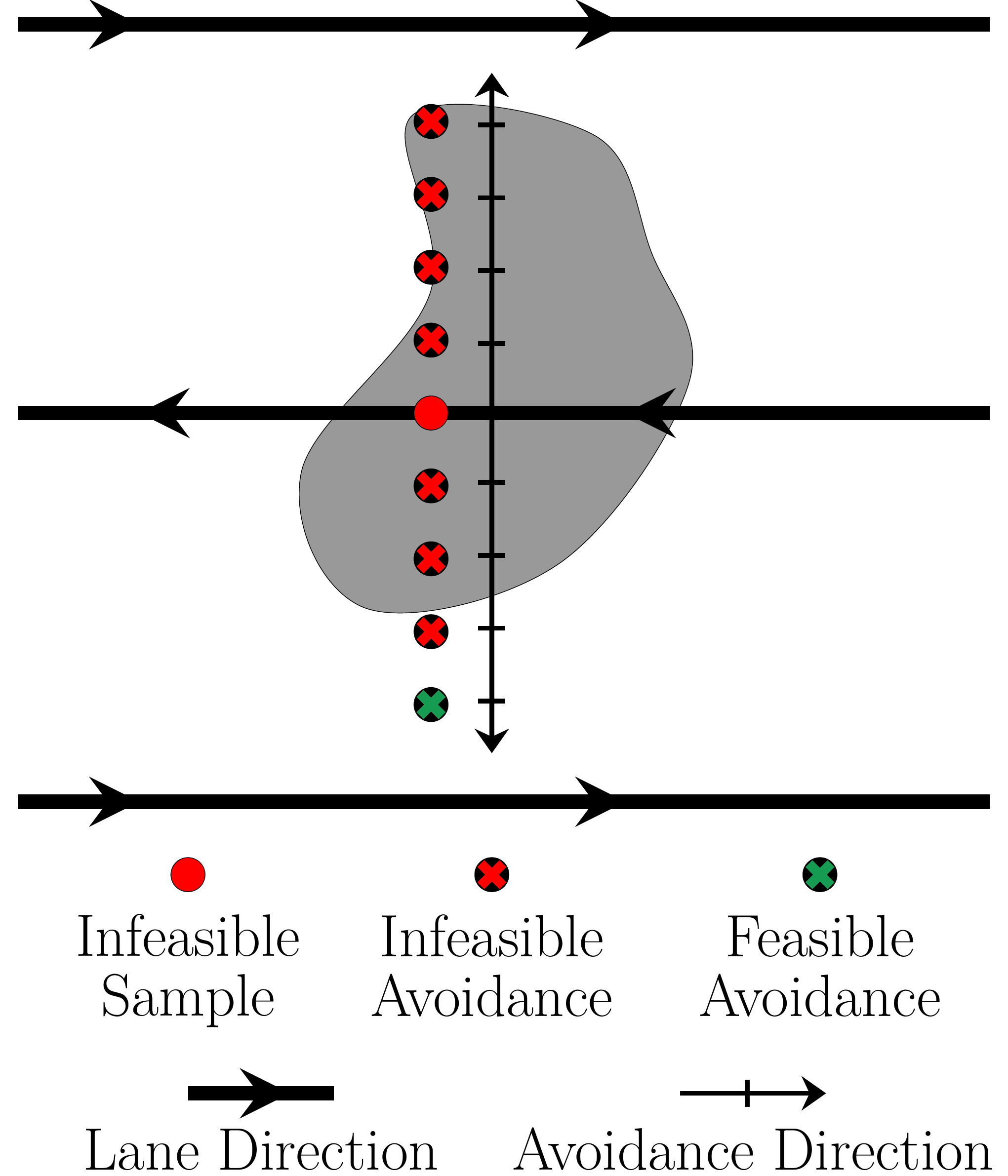}
        \label{fig:obstacle}}
        
    \caption{\textit{Left}: Illustration of the axes and angles to compute the optimal pitch angle. The grey square represents a surface cell. The red circle is the metal detector. \textit{Right}: A grey obstacle invalidates the initial sample. We incrementally generate avoidance samples in the two perpendicular directions to the lane direction until a feasible trajectory is found.}
    \label{fig:angle}
    \vspace{-4mm}
\end{figure}

At each discrete time step $t^n$, we solve a receding-horizon trajectory planning problem over the next $h$ meters of the coverage path, which can be formalized as a trajectory optimization problem starting at $t^n$, 
\begin{align}
\underset{\mathbf{R}(\cdot)}{\operatorname{min}} \hspace{4.5ex} \; & J = \int_{t^n}^{t^h} c(\dot{\mathbf{R}}(t)) \; dt & \label{eq:planning_ocp} \\
\subjectto\hspace{1ex}
& \dot{\mathbf{R}}(t) < \dot{\mathbf{R}}_\text{max}, \; & \text{(kinematic limits)} \nonumber\\
& \alpha(t) < \alpha_\text{max}, \; & \text{(surface alignment)} \nonumber\\
& |\Delta \psi_L(t) | < \Delta\psi_{L_\text{max}}, \; & \text{(forward visibility)} \nonumber\\
& \mathbf{R}(t^n) = \hat{\mathbf{R}}_{t^n}, \; & \text{(current orientation)} \nonumber\\
& \mathbf{R}(t) \in \mathit{SO}(3), & \forall t\in[t^n,t^h], \nonumber
\end{align}
where $t^h$ is the time at $h$ meter into the path and $\Delta \psi_L(t)$ is the angle between the planned heading and reference path direction as shown in Fig.~\ref{fig:corridor}. Finally, $\alpha(t)$ is the ground alignment error defined as the shortest angle between the surface normal and the $z$ axis of the detector frame shown in Fig.~\ref{fig:alignment}.

Since the platform does not roll, the alignment error for a given yaw $\psi$ is only a function of pitch $\theta$, 
\begin{equation}
\begin{aligned}
\alpha(\theta) = \cos^{-1}({}_{O}\mathbf{n} \cdot (\mathbf{R}_{OD}(\theta) \cdot {}_{D}\mathbf{z}_{D})).
\end{aligned}
\end{equation}
Following the proof in Appendix~I, we can determine the optimal pitch~$\theta^*$ that minimizes $\alpha$ for any candidate yaw, 
\begin{equation}
    \theta^*(\psi) = \tan^{-1}\left(\frac{n_x \cos(\psi) + n_y \sin(\psi)}{n_z}\right).
\end{equation}
Fixing roll and minimizing the alignment error with an optimal pitch reduces the trajectory optimization problem in \eqref{eq:planning_ocp} to just solving for the optimal yaw trajectory $\psi(\cdot)^*$ instead of the full rotation.
Finally, aiming for a smooth trajectory in terms of yaw, we define the cost function to minimize yaw changes $c(\dot{\mathbf{R}}(t)) = \dot{\psi}(t)$.

\subsection{Approximate Solution via Graph Search}
At each discrete planner iteration $t^n$, we find an approximate solution to the planning problem in \eqref{eq:planning_ocp} by incrementally constructing a state lattice \cite{Pivtoraiko2009}, and finding the minimum cost trajectory via graph search. The lattice is constructed by discretizing the 2D coverage path into reference positions. Given one discrete sample $(x_{ref}^l, y_{ref}^l)$, we calculate the 3D position using the current terrain map,
\begin{equation}
{}_{O}\mathbf{p}_{D}^l = [x_{ref}^l, y_{ref}^l, z_{ref}^l]^{\mathbf{T}} + d \cdot {}_{O}\mathbf{n}.
\end{equation}
Here $d$ is the desired distance to the surface, ${}_{O}\mathbf{n} \in \mathbb{R}^3$ is the surface normal, and $z_{ref}$ is the surface elevation.
We then sample yaw angles, as illustrated in Fig.~\ref{fig:corridor}, with a resolution of \SI{3}{\degree} to create a candidate tuple $({}_{O}\mathbf{p}_{D}^l, \psi_j^l)$. 

We attempt to connect each tuple  to all feasible tuples in the previous step $t^{l-1}$, checking that the motion satisfies the constraints in \eqref{eq:planning_ocp}, as well as checking collisions and traversability. The feasible yaw states and collision-free connections define a multitree $\mathcal{G} = \{\mathcal{V, E}\}$, as shown in Fig. \ref{fig:corridor}. Nodes $\mathcal{V}_l$ at layer $l$ represent the lattice states at the planning step $t^l$, and connect to $\mathcal{V}_{l-1}$ through edges $E(v)=\{e(v,v')\}_{v' \in N(v)}$. 
The cost along each edge is the change in yaw $c(v,v')= |\psi_{v'}-\psi_{v}|$.
The minimum cost path through the lattice is efficiently found via a breadth-first graph search. At each planning iteration, we only sample nodes for the new layer and reuse the previous tree.

\subsection{Visibility Constraint}
Our planner only allows paths through previously observed space to ensure a collision-free mission. To incentivize mapping of the area ahead, we modify the graph search to prioritize yaw candidates that enable the LiDAR to observe the \textit{next} coverage path lane. To do so, we first solve the optimization problem \eqref{eq:planning_ocp} under the additional constraint 
$$
|\Delta \psi_N(t^h)| < e_\text{max} \hspace{3ex} \text{(exploration constraint)}
$$
where $\psi_N$ is the direction that points towards the next lane, as seen in Fig.~\ref{fig:corridor}. This constraint is relaxed if no feasible solution with an exploration state was found, which can be efficiently implemented in the graph search approach by just checking the last layer in the search tree.

\subsection{Collision Avoidance}
Obstacles such as large rocks, trees, or poles may block parts of the coverage path. We identify traversable areas using a threshold on the slope of the terrain. Our algorithm pushes the path position out of the obstacle, as shown in Fig.~\ref{fig:obstacle}. For all infeasible reference positions $\mathbf{p}^{j}$ in $[t^n,t^{h}]$, we iteratively sample avoidance positions orthogonal from the coverage path until the position is feasible. Solely shifting nodes keeps the graph structure intact. Using the graph search approximation above, we then plan a trajectory over this collision-free (sub-optimal) path.

\begin{figure}[]
    \vspace{1mm}
    \centering
    \includegraphics[width=0.95\linewidth]{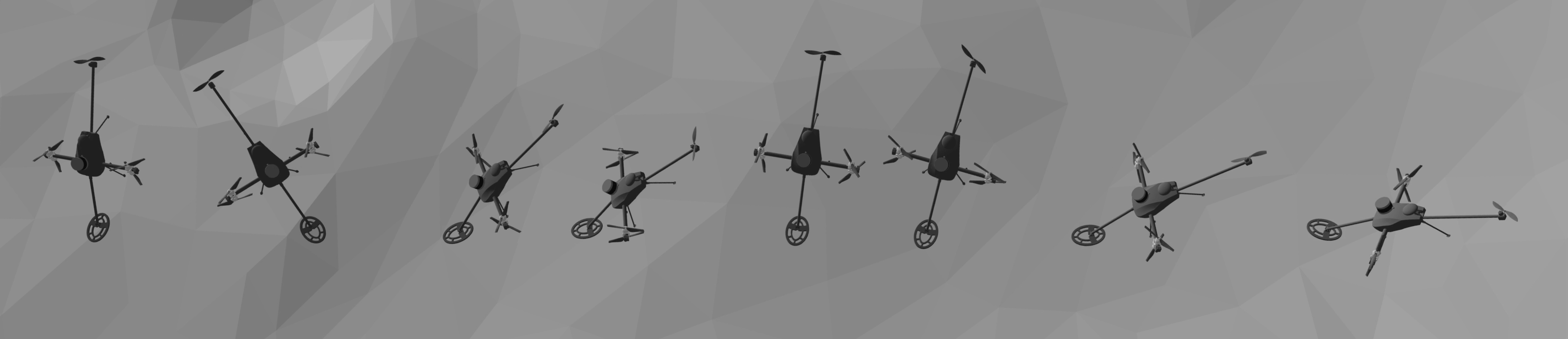}
    \includegraphics[width=0.95\linewidth]{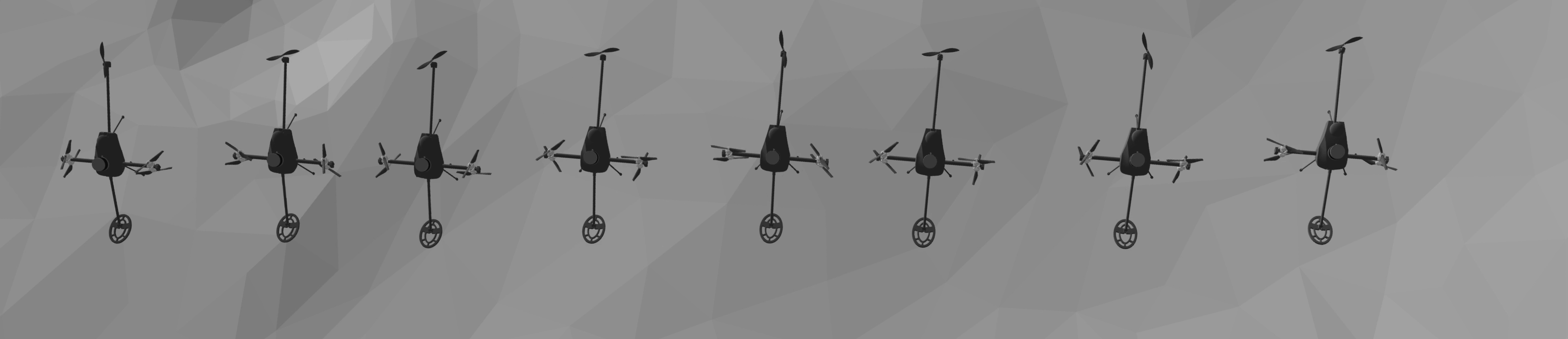}

    \vspace{-2mm}
    \caption{Irregular, mostly flat simulated terrain. \textit{Top}: The naive \textit{Aligned-1} approach results in inefficient turning. \textit{Bottom}: With the proposed planner, the system translates along the lane with only slight changes to its yaw.}
    \label{fig:sim}
    \vspace{-6mm}
\end{figure}

\vspace{-1.5mm}
\section{Experimental Results}
\vspace{-2mm}
In the following section, we first analyze the proposed planning approach quantitatively in a simulation environment. We then conduct real-world flight experiments to evaluate the sensor fusion. Finally, we validate the complete system by autonomously detecting metallic objects in the ground. We provide additional visualizations online\footnote{https://youtu.be/lv8ManXzV84}.
\vspace{-1.5mm}
\subsection{Trajectory Planner for Surface Tracking}
\vspace{-1.5mm}
To evaluate our planner, we simulate the drone using a Gazebo model with the same dimensions, \ac{LiDAR} sensor, and controller as the real platform. 
We deploy the system in a scaled-down model of a crater landscape~\cite{rheasilvia} that exhibits different types of irregular terrain that we are interested in for \ac{ERW} surveys. The 25 $\times$ 10\si{\metre\squared} survey area starts in a primarily flat but irregular terrain (see Fig. \ref{fig:sim}), followed by a curved slope, and ends in an irregular, undulated area.

In addition to the proposed planner, we simulate three baseline approaches as an ablation study. The first baseline approach (\textit{Aligned-1}) is a greedy surface tracker that only samples the two orientations that align perfectly with the surface at the next step, i.e., at $h=$ \SI{0.3}{m}. The second baseline (\textit{Aligned-6}) evaluates the optimal orientation for the next $6$ steps. The same planning horizon as our proposed planner allows evaluating the performance gain from our planning strategy independent of its ability to look ahead. The third baseline (\textit{Fixed Attitude}) is roughly equivalent to an underactuated quadcopter that ignores the slope and simply follows the terrain with a fixed height offset. We simulate this using the \si{5}~\ac{DOF} \ac{MAV} but set $\psi=\theta=0$ and mount the metal detector in the middle of the platform. Note that this will result in improved results compared to a quadcopter, as the platform can translate without pitching. All approaches use ground truth odometry. We list the used parameters in Tab.~\ref{tab:params} and present the results in Tab.~\ref{tab:sim}. To calculate alignment quality, we find the best-observed alignment $\alpha_{min}$ over all poses on the flight trajectory for each cell, and calculate the average over all cells, denoted as $\overline{\alpha_\text{min}}$.
Given the translational velocity constraint, the \textit{Fixed Attitude} method is a lower limit on how fast the area can be covered. Our approach only takes 3.6\% longer compared to this, while it covers the area in 10.6\% less time than \textit{Aligned-6} and has a much smoother trajectory in terms of yaw, as visible in Fig. \ref{fig:sim}. On average, the yaw change between subsequent target poses $\overline{\Delta \psi{t^n}}$ is decreased by \SI{23.9}{\degree} compared to \textit{Aligned-6}. As expected, our approach has a higher alignment deviation than the \textit{Aligned} approaches. However, its 95th percentile stays below the intended limit of \SI{7.5}{\degree}, which is accurate enough to enable reliable inspection. It is worth noting that extending the planning horizon from \textit{Aligned-1} to \textit{Aligned-6} only marginally improves performance. Therefore, we conclude that the increased performance in our approach mostly stems from the proposed yaw sampling strategy.
As the \textit{Fixed Attitude} baseline does not track the orientation of the surface, the metal detector is not aligned in uneven areas, resulting in $\alpha_\text{min}^\text{95th} = $ \SI{24.29}{\degree}, which is insufficient for reliable detection. Note that this result is a lower bound for a regular quadcopter, and would become even worse for more inclined areas. In contrast to our approach, all compared approaches traversed unobserved terrain multiple times. 

\begin{table}[]
\vspace{0.5mm}
\caption{Parameters used for the presented experiments}
\vspace{-3.5mm}
\begin{adjustbox}{max width=\linewidth}
\begin{tabular}{cc|cc|cc c}
& & & & & Simulation & Real-World \\
\toprule
Lane spacing & \SI{0.2}{m} & $h$ & \SI{1.8}{m} & $v_\text{max}$ & \SI{1}{m/s} & \SI{0.6}{m/s} \\ 
Sample spacing & \SI{0.3}{m} & $\alpha_\text{max}$ & \SI{7.5}{\degree} & $\omega_\text{max}$ & \SI{60}{\degree/s} & \SI{30}{\degree/s} \\ 
Grid cell size & \SI{0.15}{m} & $\Delta\mathbf{\psi}_{L_\text{max}}$ & \SI{120}{\degree} & $d$ & \SI{0.15}{m} & \SI{0.18}{m} \\ 

\bottomrule
\end{tabular}
\end{adjustbox}
\label{tab:params}
\end{table}
\vspace{-1.5mm}
\begin{table}[]
\vspace{-2mm}
\caption{Planner performance in simulation. Mean, standard deviation and 95th percentile are indicated by $\overline{\phantom{A}}$, $\sigma$ and ${}^{\text{95th}}$.}
\vspace{-3mm}
\begin{adjustbox}{max width=\linewidth}
\begin{tabular}{cccccc}
\toprule
Method & Duration [s] & $\overline{\Delta \psi_{t^n}} \pm \sigma$ [\degree] & $\overline{\Delta \theta{t^n}} \pm \sigma$ [\degree] & $\overline{\alpha_\text{min}}$ [\degree] & $\alpha_\text{min}^\text{95th}$ [\degree] \\ 
\midrule
Fixed Attitude & $2132$ & $0.0 \pm 0.0$  & $0.0 \pm 0.0$ & $11.38$ & $24.29$  \\
\hline
Aligned-1 & $2502$ & $31.28 \pm 65.06$ & $2.70 \pm 2.55$ & $1.76$ & $4.15$ \\
Aligned-6 & $2469$ & $28.07 \pm 58.15$ & $\mathbf{2.68} \pm 2.56$  & $\mathbf{1.74}$ & $\mathbf{4.14}$  \\
Proposed & $\mathbf{2208}$ & $\mathbf{4.17} \pm 17.17$ &  $2.94 \pm 3.64$ & $3.41$ & $7.25$   \\
\bottomrule
\vspace{-10mm}
\end{tabular}
\end{adjustbox}
\label{tab:sim}
\end{table}

\begin{figure}[b!]
    \vspace{-6mm}
    \centering
    \includegraphics[width=0.8\linewidth]{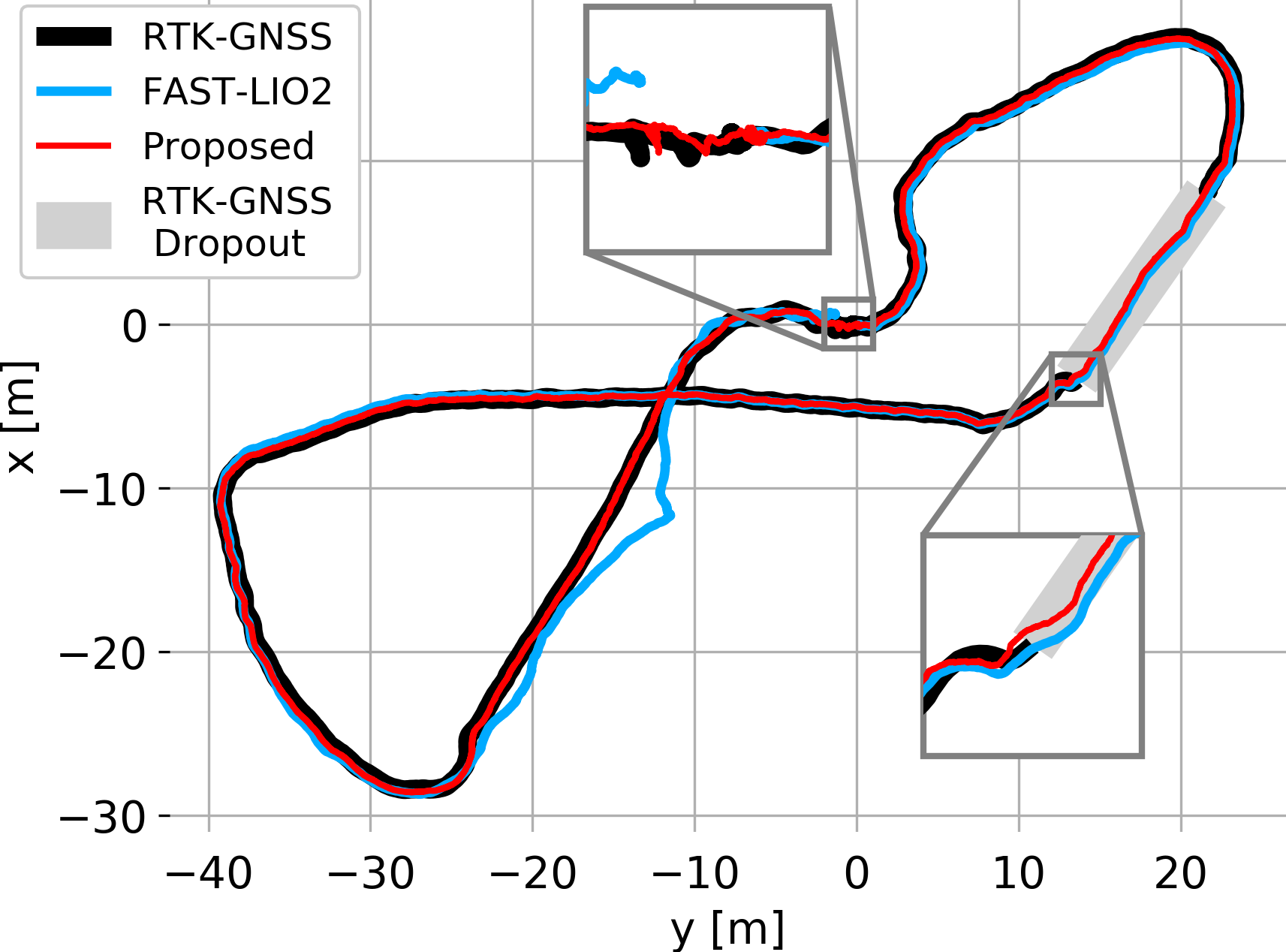}
    \vspace{-1.5mm}
    \caption{Odometry plot. \textit{Top detail}: The FAST-LIO2 trajectory is not closed due to accumulated drift. Start and end point of the proposed approach match. \textit{Bottom detail}: Our approach estimates a consistent trajectory during RTK-GNSS loss and remains smooth upon recovery.}
    \label{fig:odom}
\end{figure}

\subsection{Resilient GNSS-IMU-LiDAR Localization}
\vspace{-1mm}
To evaluate the proposed sensor fusion approach's robustness to degeneracy, we deployed our system in an outdoor environment that contains both a GNSS-denied area and an area with severe LIO drift. LIO measurements arrive at \SI{20}{\Hz}. RTK-GNSS measurements arrive at \SI{5}{\Hz}. We use a window length of \SI{5}{\second}. A pilot steered the platform in position control mode to trigger LIO and GNSS failure cases within one experiment. The controller uses the state estimation presented in Sec.~\ref{gnss_lio}. After takeoff, the extrinsics $\mathbf{T}_{OI}$ converge after approximately \SI{3.5}{m}. The drone flies through a dense tree canopy which causes RTK-GNSS loss, and later traverses flat grassland, which causes FAST-LIO2 to drift. The platform was started and landed in the same position with an accuracy of approximately \SI{0.15}{m}.

As visualized in Fig.~\ref{fig:odom}, the proposed approach retains a locally consistent state estimate while traversing through the RTK-GNSS-denied area that remains smooth upon signal recovery. FAST-LIO2 deviates \SI{4.2}{m} from the RTK-GNSS trajectory where the point cloud registration is underconstrained. Since we calculate the LIO covariance based on the geometric structure, our approach implicitly handles the degeneracy and follows the RTK-GNSS trajectory. The distance between takeoff and landing position of our approach is \SI{0.18}{m}. This is within our landing accuracy, while the final distance estimated by FAST-LIO2 is \SI{1.52}{m}. 
\vspace{-1.5mm}

\subsection{Autonomous Search for Buried Metallic Objects}
\vspace{-1mm}
\begin{figure}[t!]
    \vspace{1.3mm}
    \centering
    \includegraphics[width=0.98\linewidth]{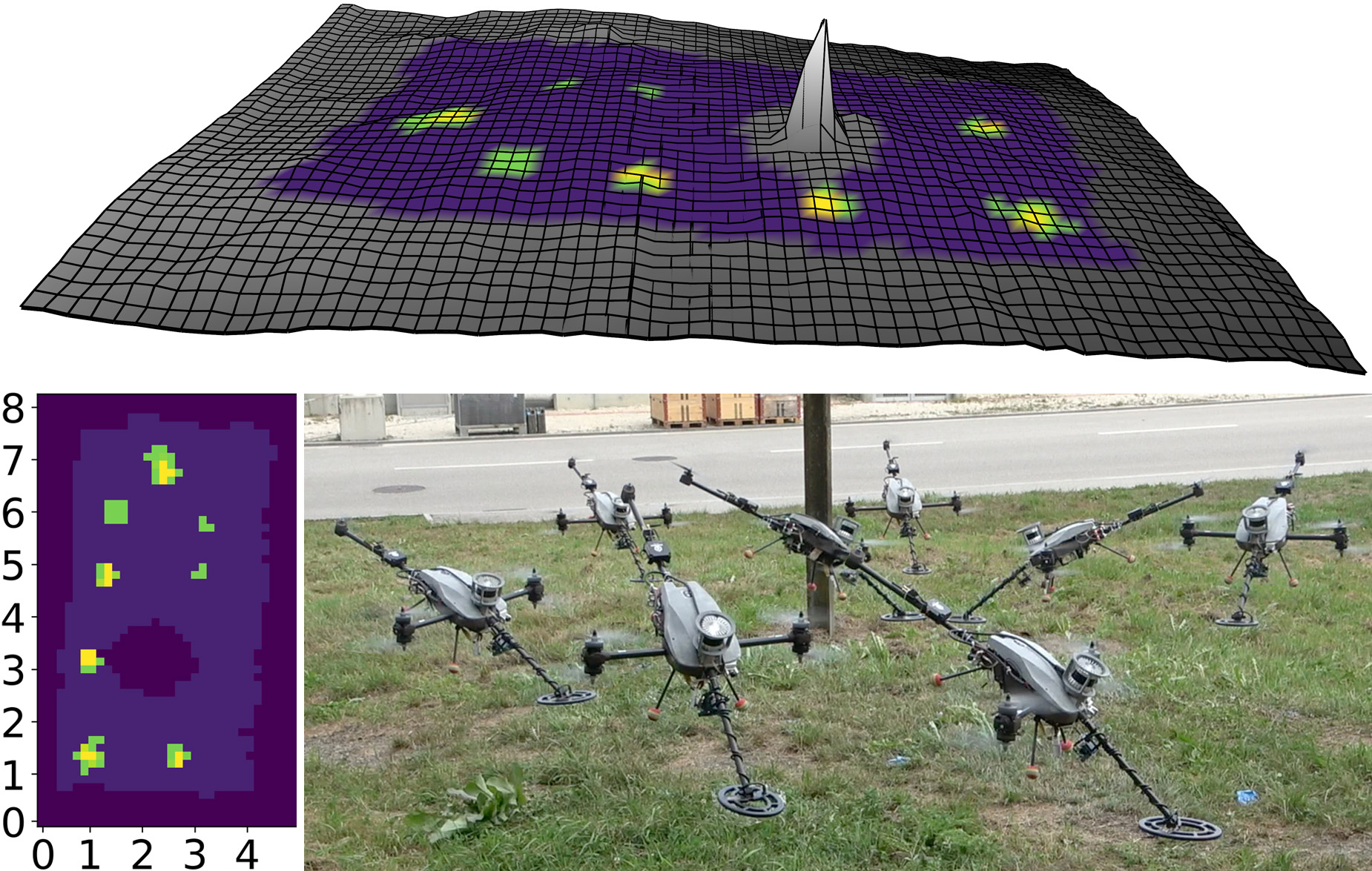}
    \vspace{-3mm}
    \caption{\textit{Top}: Metal detector signals visualized on a terrain map. Colored cells were covered by the sensor. \textit{Bottom left}: 2D detection map. \textit{Bottom right}: The system avoids the obstacle while maintaining detector alignment.}
    \label{fig:real_world}
    \vspace{-6.5mm}
\end{figure}

We evaluate the complete system in two real-world experiments. As metallic targets, we used gas cartridges with a diameter of \SI{0.1}{\metre}. They were buried in the ground to be leveled with the surface to facilitate inspection of the results. To integrate the detector signal, we identify measured cells in the grid map by ray tracing from the sensor's bounding points and center to the map. 

In the first experiment, we deploy the system in an undulated terrain covered by grass with a main slope of around \SI{20}{\degree} and steeper local inclinations along multiple axes. We placed 9 metallic targets.
The platform covered an area of 5.5 $\times$ 4\si{\metre\squared} in \SI{317}{\second}. As illustrated in Fig.~\ref{fig:undulated_heat}, the planner generates trajectories that enable the sensor head to track the surface orientation. Additionally, despite the terrain steepness, the drone avoids collisions with any part of the body. We visualize the average sensor response per cell in Fig.~\ref{fig:undulated_heat}. The 9 metallic objects are clearly visible. The lateral distance between the objects matches the actual distance of \SI{1.65}{m} by visual inspection of the 2D detection map. 

Due to the sensor's working principle, the patches with high-intensity are larger than the actual objects. One measurement represents the whole area under the sensor head. Thus, even a tiny object can cause measurements of the size of the metal detector. The sensor used in this work does not expose raw signals and performs a calibration that the manufacturer does not specify. Thus, we observed that the sensitivity increased during the experiments. For future use, we propose to use a sensor that reports raw signals to avoid inconsistent sensitivity levels.

We performed a second experiment in an environment with a main slope of around \SI{8}{\degree}, which falls off slightly to both sides. A metal pole is present in the survey area. We placed 9 metallic targets. 
The platform covered an area of 7 $\times$ 3.5\si{\metre\squared} in \SI{373}{\second}. The planner identifies the obstacle and maneuvers around it while maintaining the sensor aligned to the surface, as shown in Fig.~\ref{fig:real_world}. \SI{8}{} metallic targets are visible in the generated map. The missed target was in an area close to the pole, which the flying detector avoided. 
\vspace{-1.5mm}
\section{Conclusion}
\vspace{-1.5mm}
We presented a complete system for metal detection with a \si{5}~\ac{DOF} \ac{MAV} capable of navigating in \acs{GNSS}-denied and \ac{LIO} degenerate environments while adjusting its coverage trajectory based on terrain, visibility, and obstacles. 
The platform's maneuverability shows superior surface tracking over an underactuated \ac{MAV}.
The proposed receding-horizon planner shows good terrain perception and reduces coverage time over compared approaches.
Real-world experiments showcase that the system can autonomously survey challenging undulated terrain for ERW detection. Visualizing the metal detector response on the terrain map combined with the GNSS reference allows easy re-localization in hazardous areas for demining. In future work, the obstacle avoidance strategy, although sufficient in our experiments, could benefit from integration in a coverage re-planning layer. We also believe our resilient surface navigation holds promise for other surface tracking applications such as spraying, close-up imaging, or non-destructive testing.
\vspace{-1mm}
\section*{APPENDIX}
\vspace{-0.5mm}
\label{derivation}

\noindent We search $\theta$ that minimizes the alignment error $\alpha$ given $\psi$: 
\vspace{-4mm}
\begin{fleqn}
\setlength{\belowdisplayskip}{0pt} \setlength{\belowdisplayshortskip}{0pt}
\begin{equation}
\begin{split}
& \theta_{cand}^* = \argmin_\theta \alpha (\theta)= \argmax_\theta f(\theta)
\end{split}
\end{equation}
\setlength{\belowdisplayskip}{0.5pt} \setlength{\belowdisplayshortskip}{0.5pt}
\setlength{\abovedisplayskip}{0.5pt} \setlength{\abovedisplayshortskip}{0.5pt}
\begin{equation}
\begin{split}
& f(\theta) = {}_{O}\mathbf{n} \cdot  (\mathbf{R}_{OD}(\theta, \psi) \cdot {}_{D}\mathbf{z}_{D}) \\
& \stackrel{\text{Roll} \overset{!}{=} 0}{=} \sin(\theta)\underbrace{(n_x \cos(\psi) + n_y \sin(\psi))}_\text{a}  + \cos(\theta) n_z
\end{split}
\end{equation}
We find the critical point as: \begin{equation} \label{eqn_fd}
\resizebox{0.93\linewidth}{!}{
\setlength{\belowdisplayskip}{1.5pt} \setlength{\belowdisplayshortskip}{1.5pt}
\setlength{\abovedisplayskip}{1.5pt} \setlength{\abovedisplayshortskip}{1.5pt}
$
\begin{split}
& f(\theta_{c})' =  \cos(\theta_{c})a -   \sin(\theta_{c})n_z \overset{!}{=} 0 \\
& \Leftrightarrow \theta_{c} = \tan^{-1}(\frac{a}{n_z}) = \tan^{-1}\left(\frac{n_x \cos(\psi) + n_y \sin(\psi)}{n_z}\right)
\end{split}
$}
\end{equation}
\setlength{\belowdisplayskip}{1.5pt} \setlength{\belowdisplayshortskip}{1.5pt}
\setlength{\abovedisplayskip}{1.5pt} \setlength{\abovedisplayshortskip}{1.5pt}
\begin{equation} \label{eqn_sd}
\begin{split}
f(\theta)'' =  -\sin(\theta) a - \cos(\theta) n_z \overset{!}{<} 0 \\ \stackrel{\substack{\cos(\theta) > 0 \\ \theta \;  \in \;  (-\pi, \pi)}}{\Leftrightarrow} -\tan(\theta) a < n_z^2 
\end{split}
\end{equation}
Inserting (\ref{eqn_fd}) into (\ref{eqn_sd}) yields:
\begin{equation}
\begin{split}
-\tan(\tan^{-1}(\frac{a}{n_z})) a = -a^2/n_z < n_z \stackrel{n_z > 0}{\Leftrightarrow} -a^2<n_z^2
\end{split}
\end{equation}
\end{fleqn}
Since $f(\theta_{c})'=0$ and $f(\theta_{c})''<0$ we find $\theta^*_{cand} = \theta_{c}$

\bibliographystyle{include/IEEEtran}
\bibliography{literature.bib}

\end{document}